\documentclass[conference]{IEEEtran}
\IEEEoverridecommandlockouts
\usepackage{cite}
\usepackage{amsmath,amssymb,amsfonts}
\usepackage{algorithmic}
\usepackage{graphicx}
\usepackage{textcomp}
\usepackage{xcolor}
\usepackage{soul}
\def\BibTeX{{\rm B\kern-.05em{\sc i\kern-.025em b}\kern-.08em
    T\kern-.1667em\lower.7ex\hbox{E}\kern-.125emX}}

\begin{document}

\title{Social-Cultural Factors in the Design \\ of Technology for Hispanic People with Stroke
\thanks{This work was supported by a grant from the Vice Provost and Dean of Research (VPDoR) and the Research Development Office (RDO) at Stanford University and the National Science Foundation Graduate Research Fellowship Program.}
}

\author{
\IEEEauthorblockN{Elizabeth D. Vasquez}
\IEEEauthorblockA{\textit{Dept. of Mechanical Engineering} \\
\textit{Stanford University}\\
Stanford, USA \\
vasqueze@stanford.edu}
\and
\IEEEauthorblockN{Allison M.~Okamura}
\IEEEauthorblockA{\textit{Dept. of Mechanical Engineering} \\
\textit{Stanford University}\\
Stanford, US \\
aokamura@stanford.edu}
\and
\IEEEauthorblockN{Sean Follmer}
\IEEEauthorblockA{\textit{Dept. of Mechanical Engineering} \\
\textit{Stanford University}\\
Stanford, USA \\
sfollmer@stanford.edu}

}

\maketitle

\begin{abstract}
Stroke is a leading cause of serious, long-term disability in the United States. There exist disparities in both stroke prevalence and outcomes between people with stroke in Hispanic and Latinx communities and the general stroke population. Current stroke technology -- which aims to improve quality of life and bring people with stroke to the most functional, independent state possible -- has shown promising results for the general stroke population, but has failed to close the recovery outcome gap for underserved Hispanic and Latinx people with stroke. Previous work in health education, digital health, and HRI has improved human health outcomes by incorporating social-cultural factors, though not for stroke. In this position paper, we aim to justify accounting for unique cultural factors in stroke technology design for the Hispanic and Latinx community. We review examples of successful culturally appropriate interventions and suggest design considerations (mutually beneficial community consultation, accommodating for barriers beforehand, building on culture, and incorporating education of the family) to provide more culturally appropriate design of Hispanic and Latinx stroke technology and reduce the disparity gap.
\end{abstract}

\begin{IEEEkeywords}
stroke technology, issues of representation
\end{IEEEkeywords}


\section{Introduction}
 \textbf{The Issue of Stroke:} 
Stroke is a leading cause of serious long-term disability and death in the United States \cite{viraniHeartDiseaseStroke2021}. 3.9\% of the U.S. population older than 18 years of age is expected to have had a stroke by 2030 \cite{ovbiageleForecastingFutureStroke2013}. Stroke is caused by a failure of blood circulation in the brain from blood flow interruption as a result of either disruption in the integrity or obstruction of a blood vessel. This can cause cognitive and/or motor impairments depending on stroke location \cite{pareBasicNeuroanatomyStroke2012}. Approximately 80\% of hospitalized people with stroke have paresis at admission or during hospitalization \cite{rathoreCharacterizationIncidentStroke2002}. These residual effects make it difficult for people with stroke to care for themselves and only 25-45\% are able to return to work after their stroke \cite{gabrieleWorkLossFollowing2009, mcleanEmploymentStatusSix2007, vestlingIndicatorsReturnWork2003}. 

Medically undeserved individuals with stroke have significantly worse recovery outcomes than the general stroke population. Studies have shown that Hispanic and Latinx communities (referred to in this paper as Hispanic) in particular have higher risk factors and rates of stroke than non-Hispanic Whites \cite{viraniHeartDiseaseStroke2021, saccoStrokeIncidenceWhite1998, morgensternPersistentIschemicStroke2013, morgensternExcessStrokeMexican2004}. They also have worse functional and cognitive recovery outcomes \cite{lisabethNeurologicalFunctionalCognitive2014}. These disparities are compounded by the fact that Hispanic communities are the largest minority racial-ethnic group in the U.S. \cite{bureauDecennialCensus94171}. 

For Hispanic communities, the financial and social cost of stroke is also great. From 2005 to 2050, the total cost of stroke is projected to be \$313 billion for Hispanics, the majority of this cost coming from lost earnings \cite{brownProjectedCostsIschemic2006}. This is exacerbated by the fact that Hispanics are at a much higher risk for stroke at younger ages than non-Hispanic Whites \cite{morgensternPersistentIschemicStroke2013} -- when they are in their prime earning years. Individuals with stroke are less likely to be employed after their stroke and earn a lower hourly wage than the general population \cite{vyasLostProductivityStroke2016}. This places extreme burden on Hispanic individuals with stroke and their communities. 

 \textbf{Stroke Technology:} 
Stroke technology can ease that burden. We define stroke technology as that which aims to improve quality of life and bring people with stroke to the most functional, independent state possible. Stroke technology can take many forms, from a passive device, like a walker, to active movement therapy devices, like a rehabilitation robot. The HRI community has made many significant contributions to rehabilitation \cite{basterisTrainingModalitiesRobotmediated2014,langerTrustSociallyAssistive2019} and stroke technology with much of the progress focusing on customizing robotic rehabilitation to disability level \cite{zhangPassivityStabilityHuman2015, basterisTrainingModalitiesRobotmediated2014} or reminders and encouragement to perform rehabilitation through Socially Assistive Robots (SARs) \cite{mataricSociallyAssistiveRobotics2007, cespedesSocialHumanRobotInteraction2020}. These technologies have improved outcomes in the general stroke population. However, they are not designed for, or tested in, medically underserved Hispanic populations. Rather, they are usually developed and tested in academic medical centers with relatively affluent patients who have insurance, ample time, access to transportation, and support of a caregiver(s) \cite{polygerinosSoftRoboticGlove2015, pageEfficacyMyoelectricBracing2020, mandeljcRoboticDeviceOutofClinic2022, kimEffectsDigitalSmart2018}. Due to lack of Hispanic inclusion in design, it is unclear how effective these stroke technologies are for Hispanic communities or to what extent the disparities described above can be attributed to the noninclusive design of existing stroke technology. With the increasing prevalence of stroke, HRI insights for human-centered robotic design and DEI concerns related to stroke recovery will also grow in relevance, especially for the underserved Hispanic people.

In this paper we discuss the significance of social-cultural factors in the design of stroke technology for medically underserved Hispanic individuals with stroke in the United States and explain mechanisms for motor recovery. We review prior work in successfully implemented, culturally specific intervention for Hispanic communities in the fields of health education and digital technology as well as culturally centered work within HRI to draw insights. We propose that the design of stroke technology can more effectively support the Hispanic community by considering social-cultural factors including: mutually beneficial community consultation, accommodating for societal barriers, building on cultural norms, and educating the family.  



\section{Recovery Mechanisms}
A key component of recovery and a factor in determining ability to perform activities of daily living (ADLs) in people with stroke, is high dosage physical therapy \cite{wuLongtermEffectivenessIntensive2016a, lohseMoreBetterUsing2014a, kwakkelIntensityLegArm1999a}. Physical therapy rehabilitation can allow the brain to regain some of its lost ability to control the body through a process known as neuroplasticity \cite{lohseMoreBetterUsing2014a}. Research has shown that robot-assisted rehabilitation can aid individuals with stroke in completing more therapy and in turn improve recovery outcomes \cite{hsiehDoseResponseRelationship2012, poliRoboticTechnologiesRehabilitation2013}. Assistive technology has also improved not only movement ability, but also user satisfaction, as people with stroke use these technologies in their daily lives to perform individually meaningful tasks \cite{limEffectsAssistiveTechnologybased2020, chiuEffectTrainingOlder2004}. However, the development of stroke technology specifically for Hispanic people with stroke has not been studied, and we propose that a better understanding of how to design appropriate stroke technology for Hispanic people with stroke would lead to improved stroke technology design that reduces health outcome disparities. 

Generally, stroke technology is designed assuming that a person with stroke follows an ideal recovery journey (shown in Figure 1), when in reality the underserved Hispanic person with stroke's journey often ends after the hospital visit. When we design for the ideal, we make less accessible and applicable technology, which can exacerbate recovery disparities. Considering social-cultural factors may be key to creating a successful recovery journey for Hispanic individuals with stroke and inclusive stroke technology.

\begin{figure}[t]
\centerline{\includegraphics[scale=0.4]{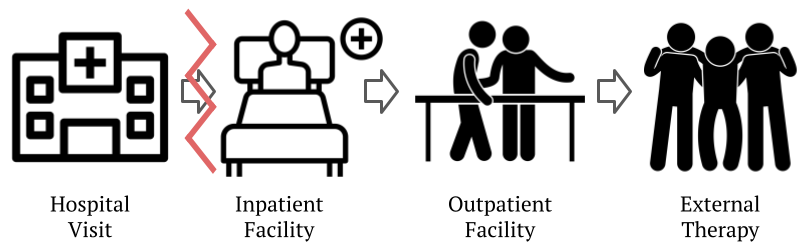}}
\caption{A simplified person with stroke’s journey to recovery. First, the patient is treated at a hospital. For the medically underserved, this journey often ends upon hospital discharge, as depicted by the red line. For the general stroke survivor, they are next discharged to an inpatient facility such as inpatient rehabilitation or a skilled nursing facility where they live and receive various therapy daily. Next, they are discharged home and receive outpatient rehabilitation a few times a week. Finally, the patient elects to participate in ongoing movement therapy at a private clinic or with a stroke support group.}
\label{fig:ideal_journey}
\end{figure}


\section{Successful Culturally Specific Interventions}
 
 In this section, we summarize the methods and takeaways from culturally specific Hispanic interventions in health education and digital health. We then present relevant culturally oriented HRI work. 
 
 Prior health education, culturally tailored programs for Hispanic communities have focused on improving knowledge, self-management, and reducing risk factors of disease  \cite{benderCulturallyAppropriateIntervention2013, brunkCulturallyAppropriateSelfManagement2017, evans-hudnallImprovingSecondaryStroke2014a, teufel-shoneDevelopingAdaptingFamilybased2004}. Hispanic communities often face situational and structural challenges such as cost, lack of insurance, lack of transportation, immigration stigmatization, and effects of discrimination \cite{masonHispanicCaregiverPerceptions2015, dominguezVitalSignsLeading2015, teufel-shoneDevelopingAdaptingFamilybased2004, wolfeTransportationBarriersHealth2020, saucedoMexicansImmigrantsCultural2012, delatorreBuriedVoicesMexican2013, casanovaStigmatizationResilienceFemale2012, cuttsCommunityHealthAsset2016a} which can result in behavior that leads to worse health outcomes. To address these issues, education can facilitate healthier behaviors from patients and address core challenges that cause unhealthy behaviors to create an environment conducive to maintaining those new actions \cite{pickIntegratingInterventionTheory20030728}. The Hispanic culture has many aspects to facilitate this process. One study used  \emph{familialism} -- ``the strong identification and attachment of individuals with their nuclear and extended families as well as the presence of strong feelings of loyalty, reciprocity, and solidarity among members of the same family'' \cite{marinDefiningCulturallyAppropriate1993} -- to help Hispanic Type II diabetics control their disease by educating and setting health-behavior goals at the family level \cite{teufel-shoneDevelopingAdaptingFamilybased2004}. Incorporating religiosity and collectivism has also yielded success \cite{evans-hudnallImprovingSecondaryStroke2014a, caballeroUnderstandingHispanicLatino2011}. 


 These public education interventions used human-centered design and incorporated community health workers and organizations who know the language and culture to help recruit, facilitate, disseminate, and upscale the research in a community-sustainable way \cite{pickIntegratingInterventionTheory20030728, benderCulturallyAppropriateIntervention2013}. Interventions focused on providing patients and their caregivers with communication and decision making skills to effectively problem solve and improve their self-esteem and self-efficacy \cite{pickIntegratingInterventionTheory20030728}.

  Culturally appropriate digital health has also found success in Hispanic communities \cite{pekmezarisAdaptingHomeTelemonitoring2020, pekmezarisTelehealthDeliveredPulmonaryRehabilitation2020, leeDesigningEffectiveEHealth2022, jacobsDevelopmentCulturallyAppropriate2014}. Many used a Community-Based Participatory Research (CBPR) method in which stakeholders help make decisions, define research questions, collect and analyze data, interpret findings, and disseminate research. A Community Advisory Board (CAB) -- consisting of community stakeholders -- provides insight on community challenges, barriers, and suggestions for adapting a technology \cite{pekmezarisAdaptingHomeTelemonitoring2020, pekmezarisTelehealthDeliveredPulmonaryRehabilitation2020}. Other approaches use qualitative research techniques, such as surveys, to determine challenges, barriers, and culturally relevant adaptations \cite{jacobsDevelopmentCulturallyAppropriate2014}. A key difference between CBPR and these other methods is that CBPR promotes trust and shared power. This means that the data is jointly owned by the researchers and the community members, co-learning occurs (academic and community partners learn from each other), community capacity is built (researchers train community members in research), and a long-term commitment to reduce disparities is made \cite{flickerEthicalDilemmasCommunityBased2007, mikesellEthicalCommunityengagedResearch2013, wallersteinUsingCommunityBasedParticipatory2006}. Key points to consider are dealing with dissemination of sensitive or unflattering data and caution towards inadvertently causing conflict between community members \cite{wallersteinUsingCommunityBasedParticipatory2006, mikesellEthicalCommunityengagedResearch2013, flickerEthicalDilemmasCommunityBased2007}.
  
 Researchers using a community-based design approach have found that to make culturally appropriate adaptations to technology for Hispanic populations, intervention should be comfortable (ergonomic, simple to understand directions/use, use appropriate language level and examples from familiar aspects of culture) and account for the challenges of the culture (e.g. long work hours necessities flexible appointment booking) \cite{pekmezarisAdaptingHomeTelemonitoring2020, pekmezarisTelehealthDeliveredPulmonaryRehabilitation2020, leeDesigningEffectiveEHealth2022, jacobsDevelopmentCulturallyAppropriate2014}.
   
  
  HRI research has also incorporated culture, but it is generally defined on a nation-based rather than ethnic-based level. Incorporating culturally aligned cues and behaviors in robots has resulted in better interaction \cite{wangWhenRomeRole2010, papadopoulosCARESSESRandomisedControlled2021, limSocialRobotsGlobal2021}.
  In the case of SARs, which hold great promise for rehabilitation \cite{langerEmergingRolesSocial2021}, recent research suggests that including culturally competent AI could lead to improved psychological well-being in the elderly \cite{papadopoulosCARESSESRandomisedControlled2021}. 
  However, to the best of our knowledge, no work in HRI has looked at how interaction with robots for stroke technology should change to engage with ethnically Hispanic groups. 
  

\section{Design Considerations}
 
 Building on methods and insights from health education and technology communities, we propose design considerations for Hispanic stroke technology development that ethically leverages aspects of the culture and community.
 
 

 \textbf{Building a Two-way Street of Community Research:}
 Researchers in HRI have and must continue to include the communities they want to work with. For Hispanic people with stroke, researchers should incorporate a community partner such as a local stroke support groups, community non-profits, rehabilitation facilities, local churches, etc. Researchers are encouraged to include community members throughout the entire research process, which is not always done in HRI. To our knowledge, no community-based HRI research has looked at Hispanic populations in particular, but some have sought to understand the implications of other communities \cite{moharanaRobotsJoyRobots2019}. Two key factors this prior HRI work has not fully considered are (1) the effect of power balance and (2) reciprocity. For power balance, we encourage researchers to verify insights gained from community interactions to ensure they are not misrepresentations or invented insights about a community experience that researchers do not live. Particularly for Hispanic groups who may prefer another language, it is important that this verification happen in appropriate language. Towards reciprocity, we encourage researchers to give back to the community, be that through a commitment to community capacity building (mentioned earlier in our description of digital health interventions) or through providing a platform for the community to have a voice. Examples are a research-team-led educational seminar to teach the community about (1) controlling stroke risk factors or (2) research methods used in the study. In this way researchers can build trust, verify the applicability of research, and reduce disparities.
 

 \textbf{Accommodate for Societal Barriers Beforehand:}
  Researchers should recognize personal barriers that a Hispanic person with stroke experiences and ensure that stroke technology and participating in research does not exacerbate them. For example, a Hispanic medically underserved individual might have difficulty finding transportation to health services and family members might work long hours \cite{evans-hudnallImprovingSecondaryStroke2014a, garciaEngagingIntergenerationalHispanics2019}, so a technology that requires a clinic visit may result in less adherence than an at-home one. Practically, technologies should not require an expert to use them. If the device was an exoskeleton, it should be easily donnable and doffable. If additional individual are required for use, the technology could facilitate connecting the user with self-nominated volunteers in the community for help if the immediate family is busy working. Similarly, requiring a Hispanic person with stroke to participate in research at set hours may be burdensome on the family. Offering to go to them and being flexible with timing are essential. Looking at the healthcare experience of medically underserved Hispanics shows that no or lower quality insurance, a lack of Spanish-speaking providers or providers with cultural awareness, minimal non-urgent or non-emergent medical care, difficulty paying out of pocket, and negative family pressures can all play an adverse role in the healthcare experience and in the use of any potential stroke technology \cite{evans-hudnallImprovingSecondaryStroke2014a, teufel-shoneDevelopingAdaptingFamilybased2004, benderCulturallyAppropriateIntervention2013, caballeroUnderstandingHispanicLatino2011, leeDesigningEffectiveEHealth2022}.

 \textbf{Build on the Culture:}
 Researchers should build on Hispanic culture to motivate use and usefulness of stroke technology. For example, research has shown that Hispanic wives -- who often act as caregivers in Hispanic families -- experience higher levels of caregiver burden than non-Hispanic white wives \cite{garciaCaregivingContextEthnicity}. This burden can lead to depressive symptoms in caregivers which have been correlated to lower health-related quality of life for the person with stroke \cite{kauhanenPoststrokeDepressionCorrelates1999}. Research also found that Hispanics were 0.78 times less likely to use assistive devices than non-Hispanic Whites \cite{resnikRacialEthnicDifferences2006}. A possible explanation is that Hispanic families feel culturally obligated to provide personal (human) caregiving, and thus the person with stroke is less likely to use devices. Researchers may be able to leverage these components -- cultural need to provide family care and the correlation between person with stroke and caregiver's health -- 
 by identifying caregiver tasks and using stroke technology to facilitate the person with stroke to assist the caregiver -- even if only symbolically -- thus fulfiling familial need to help through group participation of both the family and the person with stroke. Caregiver burden would be lifted either in the actual task or in the perception of it, and the person with stroke would benefit from feeling that they are contributing to their own care. 
 
As an example, providing emotional support to a person with stroke is one of the most difficult and time consuming tasks for a caregiver \cite{bakasTimeDifficultyTasks2004}. To address this, a stroke technology could monitor the emotional state of both the person with stroke and the caregiver. If needed, the technology could provide mental wellness strategies, help articulate emotional needs to others, or connect them to external human support. Through these features, the technology facilitates family support for both the caregiver and the person with stroke and in doing so reduces caregiver burden of being solely responsible for the person with stroke's emotional state, improves well being of the caregiver, and empowers the person with stroke. In this way, we can use cultural factors to support the family unit in a way that parallels previous health education work \cite{teufel-shoneDevelopingAdaptingFamilybased2004}. 

 \textbf{Teach a Family to Fish:} 
 Researchers should include features in stroke technology that help people with  stroke -- and their families -- understand and control their disease state. In many Hispanic families due to expectations of gender roles, women have the most knowledge about healthcare \cite{caballeroUnderstandingHispanicLatino2011}. However, family history of stroke predisposes people for a stroke \cite{pareBasicNeuroanatomyStroke2012}, so it is essential that all members of the family of a person with stroke understand risk factors and how to control them. There is a need to share health knowledge and educate the family on stroke risk factors. Incorporating features that facilitate family-based education and control of risk factors could prevent further strokes, thus reducing recovery outcome disparities. Previous work in diabetes Type II education found improved disease knowledge using a family-based approach \cite{teufel-shoneDevelopingAdaptingFamilybased2004}. One potential implementation is a metric that users and families can observe and change together. For example, a technology that shows how exercise time reduces blood pressure (a risk factor in stroke) and facilitates family exercise to reduce risk factors in the family as a whole. This sort of observable metric and family-unit exercise has been successful in preventing obesity within the Hispanic community \cite{benderCulturallyAppropriateIntervention2013} and could be integrated into stroke technology. Whereas HRI work often focuses on the individual user \cite{georgiouApplyingParticipatoryDesign2020a, mataricSociallyAssistiveRobotics2007, efthimiouUserCenteredDesign2019}, we can consider the family as a user.


\section{Challenges \& Barriers to Inclusive Design}
Inclusive design of stroke technology for Hispanic communities has distinct challenges that we review below.

\textbf{Building Trust:}
First, challenges can arise from reservations that Hispanic people have towards research participation \cite{leeDesigningEffectiveEHealth2022, wallersteinUsingCommunityBasedParticipatory2006} and fears based on immigration status \cite{teufel-shoneDevelopingAdaptingFamilybased2004}. Some of this hesitancy stems from historically experienced ill treatment from researchers who have taken from these communities without giving back. We must work to earn back the trust that we have lost. During recruitment, some researchers have found it helpful to give a face to a name by putting a picture of the their face on recruitment material and visiting community meetings in person \cite{leeDesigningEffectiveEHealth2022}. In the early stage of studies, researchers have built trust by acknowledging personal and institutional histories of misconduct and offering a platform for frank truth-telling about these traumas. During the study, researchers should use ethical practices \cite{flickerEthicalDilemmasCommunityBased2007, mikesellEthicalCommunityengagedResearch2013, wallersteinUsingCommunityBasedParticipatory2006}, be present, and listen and value the expertise and lived experience of all community members \cite{christopherBuildingMaintainingTrust2008, cuttsCommunityHealthAsset2016a}. 



\textbf{The More the Merrier:}
When it comes to using HRI to improve health issues, there are interactions beyond the system and people -- there exist larger systemic issues. Sociology can help us understand the complex intricacies of institutionalized racism and how the experience of a Hispanic person with stroke might be shaped by the surrounding cultural and social structures. This can give us greater insight as we seek to learn from and be trusted by people with stroke and their families during a community-based design process. Public health specialists can advise on approaches to medical care in a cultural-social content. Stroke neurologists have expertise in the science and treatment of stroke and thus are essential to make sure that the technology we build is medically effective. 



\textbf{Kindness is Key:}
Finally, it is important to remember when we speak to community members and stakeholders that they are people whose lives have been disrupted by a serious disease. Individuals with stroke may struggle on a daily basis with their mental health and coming to terms with their post-stroke body \cite{torregosaDealingStrokePerspectives2018, kauhanenPoststrokeDepressionCorrelates1999}. While we focus here on stroke technology for addressing sensorimotor deficits, people with stroke may also suffer from mild to severe cognitive deficits \cite{delavaranCognitiveFunctionStroke2017a}. It is a privilege to hear their stories and experiences. It may take more effort and time for a person with stroke to express themselves, but it is important that we listen.

\section{Conclusion}
Hispanic and Latinx people with stroke face worse recovery outcomes than the general stroke population, yet they are largely unrepresented in current stroke technology design and evaluation. They face many barriers to adequate health care and recovery channels, but through culturally appropriate design we may be able to build on the strengths of the Hispanic community to create more effective stroke technology capable of improving recovery outcomes. 

In this paper, we identified the extent of the stroke outcome disparity between Hispanic people with stroke and the general stroke population. We explained the mechanisms by which recovery from stroke is possible and how stroke technology can help. We followed this with a discussion of key findings from successful culturally specific educational and technology-based, health interventions for the Hispanic community. Synthesizing this, we proposed additional design considerations to optimize the design of stroke technology for the Hispanic community of people with stroke in the United States. Finally, we presented challenges to designing for medically underserved Hispanic people with stroke. Though the challenges and considerations presented here are based on U.S.\ populations, they may also apply elsewhere.

Society on the whole has benefited from the work done in the fields of robotics and HRI, but those benefits have seldom reached the Hispanic underserved. Hispanic culture is incredibly vibrant, and we as researchers have the opportunity to develop more effective designs by keeping that culture in mind. We challenge ourselves and the HRI research community to incorporate the design considerations provided in this paper to yield successful human interactions between underserved Hispanic people with stroke and rehabilitation robots as well as other forms of stroke technology. In doing so, we may be able to reduce the disparities experienced by Hispanic individuals with stroke. 



\section*{Acknowledgements}

The authors thank Dr.~Caitlyn Seim, Dr.~Maarten Lansberg, Dr.~Marion Buckwalter, and Dr.~Asad Asad for their insights and collaboration on our ongoing research to develop technology for medically underserved Hispanic people with stroke.


\bibliography{references}

\end{document}